\documentclass[letterpaper]{article} 
\usepackage[]{aaai23}  
\usepackage{times}  
\usepackage{helvet}  
\usepackage{courier}  
\usepackage[hyphens]{url}  
\usepackage{graphicx} 
\usepackage{amsmath,amssymb,amsthm}
\usepackage{natbib}  
\usepackage{caption} 
\usepackage{subcaption}
\usepackage{xcolor}

\usepackage[linesnumbered,ruled,vlined,noend]{algorithm2e}

\usepackage{newfloat}
\usepackage{listings}
\DeclareCaptionStyle{ruled}{labelfont=normalfont,labelsep=colon,strut=off} 
 \nocopyright

\title{On Guiding Search in HTN Temporal Planning with non Temporal Heuristics}
\author{
    Nicolas Cavrel, Damien Pellier, Humbert Fiorino
}
\affiliations{
    \textsuperscript{}Univ. Grenoble Alpes - LIG \\
        Grenoble, France \\
\{nicolas.cavrel, damien.pellier, humber.fiorino\}@univ-grenoble-alpes.fr

}

\newcommand{\name}{\text{\it name}}
\newcommand{\pre}{\text{\it precond}}

\newcommand{\effect}{\text{\it effect}}
\newcommand{\add}{\text{\it effect}^{+}}
\newcommand{\del}{\text{\it effect}^{-}}

\newcommand{\tstart}{\text{\it start}}
\newcommand{\tend}{\text{\it end}}
\newcommand{\tinv}{\text{\it inv}}
\newcommand{\task}{\text{\it task}}

\newcommand{\subtasks}{\text{\it subtasks}}
\newcommand{\constraints}{\text{\it constraints}}

\begin{document}

\maketitle

\begin{abstract}
The Hierarchical Task Network (HTN) formalism is used to express a wide variety of planning problems as task decompositions, and many techniques have been proposed to solve them. However, few works have been done on temporal HTN. This is partly due to the lack of a formal and consensual definition of what a temporal hierarchical planning problem is as well as the difficulty to develop heuristics in this context. In response to these inconveniences, we propose in this paper a new general POCL (Partial Order Causal Link) approach to represent and solve a temporal HTN problem by using existing heuristics developed to solve non temporal problems. We show experimentally that this approach is performant and can outperform the existing ones.
\end{abstract}

\section{Introduction}
\label{introduction}
Among planning formalisms, Hierarchical Task Network (HTN) planning is one of the most expressive. In addition to classical STRIPS actions, HTN allows to express complex abstract tasks in the form of decompositions into subtasks and their ordering constraints. HTN has been used in a wide variety of applications \cite{europa, htnRobotics, htnTaskAuction}. However, despite the pioneering work of \cite{ixtet,shop2,goldman06}, few approaches have been proposed to deal with time in HTN planning.

Temporal HTN approaches can be classified according to the expressiveness of the solution plans they can generate in the sense of Cushing's classification of temporal problems \cite{cushing}. This classification defines three categories of temporal planning problems. The first category contains the temporal problems whose solutions are solely sequential solution plans (non-concurrent solution plans). The second one contains the temporal problems whose solution plans can possibly be concurrent, but for which there exists a sequential solution plan (possibly concurrent solution). Finally, the third one includes the temporal problems for which the only existing solution plans are necessarily concurrent (necessary concurrent solution).

Most of the approaches are able to solve only temporal problems of the first two Cushing's categories. Among these approaches, we distinguish state-space approaches \cite{shop2, siadex}. These approaches deal with temporal actions as classical ones by either preprocessing the temporal actions into a sequence of classical tasks \cite{shop2}, or by compiling temporal actions into classical actions. The latter approach is also widely used in STRIPS planning \cite{fox03,jimenez15}. Approaches converting temporal problems into classical ones can benefit from the classical search heuristics and algorithms. At the same time, several plan-space approaches have been proposed \cite{Younes03,bechon14, fape}. They aim at refining an initial HTN into a solution by refining tasks and building causal relations between them, and use \textit{STN} (Simple Temporal Networks) \cite{dechter91} to deal with temporal constraints. These approaches are relatively efficient but suffer from their lack of flexibility, especially when dealing with real problems, where tasks are necessarily concurrent.

To our best knowledge, the only approach capable of producing expressive plans of the third Cushing's category is an algorithm in plan-spaces that handles constraints using \textit{Chronicles} \cite{fape}. Chronicle approaches keep tracks of the value of each proposition with regards to time into a \textit{timeline}, then try to find non conflicting timelines for every proposition. These approaches are very expressive but suffer from their lack of efficiency due to the lack of informative heuristics.

In this paper, we propose a new planning approach for hierarchical temporal planning able to solve planning problems for the two first  Cushing's categories, called HTEP ({\it Hierarchical Temporal Event Planner}). HTEP is based on POCL ({\it Partial Ordered Causal Link}) \cite{bercher_hybrid}. The particularity of our approach is to be both flexible (HTP produces partial temporal solution plans) and to be able to exploit the heuristics developed for non temporal HTN planning for plan and flaw selection. To do so, HTEP starts by compiling a temporal problem into a non temporal one by refining abstract tasks and durative tasks into instantaneous non temporal actions. Then, it tries to find a solution plan by relaxing the duration constraints on tasks and by checking only their consistency. The relaxed non-temporal problem is expressed as a partial plan in the POCL semantics. Hence classical POCL search algorithms and heuristics can be used to solve it. Finally, if a solution to the relaxed problem is found, HTEP tries to find timestamp assignments matching the temporal constraints by using a simple CSP solver.

The paper is organized as follows. Section 1 introduces the temporal planning formalism. In the second section we present our planner, from the relaxed problem to the search algorithm solving it, with the heuristic used to guide it. Finally, the third section shows the performance results of HTEP.

\section{Problem Statement}
\label{problemStatement}

In this section we propose a formalization of a temporal HTN planning problem and its solution. The notations are based on \cite{holler20} and \cite{abdulaziz22} to deal with time.

\subsection{Action, Methods, Tasks and Plan}

A key concept in HTN planning and a fortiori in temporal HTN planning is the concept of task. Each task is given by a name and a list of parameters. We distinguish three kinds of tasks: the snap actions, the durative actions and the abstract tasks. Unlike snap actions that do change the state of the world, {\it durative tasks} and {\it abstract tasks} do not. They are names referring to other tasks (either snap, durative or abstract) that must be achieved with respect to some constraints. We consider every task has a start and an end time point. We refer to the start and the end time points of a task $t$ with temporal variables denoted respectively $v^{s}_{t}$ and $v^{e}_{t}$. Since a snap action $t$ is instantaneous, $v^{s}_{t}  = v^{e}_{t}$, thus we will simply refer to the time point of the snap action $t$ as $v_{t}$.

The durative actions and the abstract tasks can be refined respectively by applying {\em snap actions} and {\em methods} defined below.

A {\it snap action} $a$ is nearly an action in the sense of classical planning, i.e., a tuple $(\name(a), \pre(a),$ $\effect(a))$. $\name(a)$ is the name of $a$. The preconditions $\pre(a)$ and effects $\effect(a)$ are sets of ground predicates. Let $v_a$ be the time point at which $a$ is supposed to be executed. $a$ is executable if $\pre(a)$ hold strictly before $v_a$. As in classical planning, the execution of $a$ produces the effects $\effect(a)$ such that $\effect(a) = \add(a) \cup \del(a)$ and $\add(a) \cap \del(a) = \emptyset$ where $\add(a)$ and $\del(a)$ are conjunctions of predicates, respectively true and false after the execution of $a$. Finally, we say that a snap action $a$ refines a task $t$ if $t = \name(a)$.

A {\em durative action} $a$ is a tuple $(\name(a), \tstart(a), \tend(a),$ $\tinv(a), d)$: $\name(a)$ is the name of $a$, $\tstart(a)$ and $\tend(a)$ are snap actions ; $\tinv(a)$ is a set of ground predicates that must hold after the execution of $start(a)$ and until the beginning of $end(a)$, i.e., on the interval $]v^{s}_a, v^{e}_{a}[$ and $d = v^{e}_a - v^{s}_a$ is the duration of $a$. We assume as PDDL~2.1 \cite{fox03} that $v^{s}_a < v^{e}_a$ is true. Therefore the duration of $a$ is a strictly positive number. Similarly to a snap action, a durative action refines a task $t$ if $t = \name(a)$.

An {\em abstract task} must be decomposed into durative actions in order to be performed. The several ways of decomposing an abstract task are described through \textit{methods}. Even though we cannot set a duration for an abstract task in the general case, the start and end point of an abstract can still be subject to temporal constraints.

A {\em method} $m$ is a tuple $(\name(m), \task(m), \subtasks(m), \alpha,$ $\constraints(m))$, where $\name(m)$ is the name of the method, $\task(m)$ is the task refined by the method, $\subtasks(m)$ the set of tasks symbols (possibly empty) which refines $\task(m)$, $\alpha: \subtasks(m) \mapsto {\cal T}$ maps the task symbols to a set of task names and $\constraints(m)$ is a set of temporal ordering constraints over $\subtasks(m)$. Temporal ordering constraints are defined over the time variable start or the end of the subtasks $\subtasks(m)$ of $m$. The possible qualitative temporal ordering constraints are those from the classical point algebra \cite{broxvall03}: $<$, $\leq$, $>$, $\geq$, $=$ and $\neq$. For instance, the temporal ordering constraint $v^{s}_{t_1} < v^{e}_{t_2}$ expresses that the start of the task $t_1$ must occur strictly before the end of $t_2$. A method $m$ refines a task $t$ if $t = task(m)$. Note, that consistency checking of such a set of constraints  $C$  can be refined by computing strongly connected components of the constraint graph associated in polynomial time $O(|C|)$.

A {\em partial temporal plan} $\pi$ is a tuple $(T, \alpha, {\cal{C}}, {\cal{L}})$ where:
\begin{itemize}
\item $T$ is a set of task symbols.
\item $\alpha: T \mapsto {\cal T}$ a mapping from task symbols to task names.
\item ${\cal{C}}$ is a set temporal ordering constraints over the tasks symbols in $T$. The constraints are like those used in methods.
\item ${\cal{L}}$ is a set of causal links of the form $\langle t_i \overset{p}{\rightarrow} t_j \rangle$ with $t_i$ and $t_j$ two snap actions in $T$ such as $(t_i < t_j) \in {\cal{C}}$ and $p \in \effect(\alpha(t_i))$ and $p \in \pre(\alpha(t_j))$ (classical causal link definition in POCL).
\end{itemize}


A snap task $t_k$ in a partial temporal plan $\pi$ is a {\em threat} on a causal link $\langle t_i \overset{p}{\rightarrow} t_j \rangle$ if and only if (1) $t_k$ has an effect $\neg p$ and (2) the ordering constraints $(t_i < t_k)$ and $(t_k < t_j)$ are consistent with ${\cal{C}}$ if $t_i < t_j$.

A {\em flaw} in a partial temporal plan $\pi = ({\cal{T}}, {\cal{C}}, {\cal{L}})$ is either (1) an open precondition, i.e., a precondition or a postcondition of task $t \in {\cal{T}}$ not supported by a causal link or (2) a threat, i.e., a task that may interfere with a causal link or (3) a task $t \in {\cal{T}}$ that is not a snap task.

 \subsection{Temporal HTN Planning Problem and Solution}
 \label{THTN}

A {\em temporal HTN planning problem} $\cal{P}$ is a tuple $(L, {\cal{T}}, {\cal{A}}, {\cal{M}}, {s_0}, \pi_0, g)$, where $L$ is a finite set of logical propositions, $\cal{T}$ is a set of tasks, ${\cal{A}}$ is a set of durative actions and $\cal{M}$ is the set of methods, $s_0 \subseteq L$ is the initial state in the set of states ${\cal{S}}$, $\pi_0$ is the initial partial temporal plan, and $g \subseteq L$ is a of ground predicates describing the goal.

The solution of a temporal HTN planning problem is a partial temporal plan $\pi$ obtained by refining an initial partial plan $\pi_0$ as in POCL planning built into snap tasks by applying methods and durative actions. Formally, a partial temporal plan $\pi$ is solution of a planning problem ${\cal{P}} = (L, {\cal{T}}, {\cal{A}}, {\cal{M}}, {s_0}, \pi_0, g)$ if and only if:
\begin{enumerate}
    \item $\pi$ is a refinement of the initial partial temporal plan $\pi_0$:  $\pi_0$ contains two special snap task: $t_0$ with no precondition but with $s_0$ as effects and $t_\infty$ with the goal $g$ as precondition but no effects and $v^{e}_{t_0} < v^{s}_{t_\infty}$ in ${\cal{C}}$.
    \item $\pi$ needs to be executable in the initial state $s_0$. Thus,
    \begin{enumerate}
        \item all tasks in $\pi$ are snap tasks,
        \item $\pi$ has no flaws, i.e., no open precondition and no causal threats,
        \item for all $t$ in $\pi$, $v_{t}$ is assigned and satisfies the temporal constraints of $\pi$.  
    \end{enumerate}
\end{enumerate}

It remains to define how to refine a partial temporal plan into a plan containing only snap actions by using methods and durative actions.

First, consider the case of the method refinement. Let $m = (\name(m), \task(m), \subtasks(m), \alpha,$ $\constraints(m))$ be a method that refines a task $t$ and plan $\pi_1 = (T_1, \alpha_1, {\cal{C}}_1, {\cal{L}}_1)$ a plan such that $t \in {\cal{T}}_1$. Then, $m$ refines $\pi_1$ into a plan $\pi_2 = (T_2, \alpha_2, {\cal{C}}_2, {\cal{L}}_2)$ and
\begin{equation*}
\label{method-refinement}
\begin{aligned}
T_2 = & \ (T_1 - \{t\}) \cup \subtasks(m) \\
\alpha_2 = & \{(t', \alpha_1(t'))|\ t' \in T_1 \backslash \{t\}\} \cup \alpha \\
{\cal{C}}_{2} = & \ {\cal{C}}_{1} \cup \constraints(m)  \\
        & \cup \{ c \ | \ \forall u \in \subtasks(m) \ v^{s}_{t} \leq v^{s}_{u} \}  \\
        & \cup \{ c \ | \ \forall u \in \subtasks(m) \ v^{e}_{t} \geq v^{e}_{u} \}  \\
        & \cup \{ v^{s}_{t} \leq v^{e}_{t} \}  \\
{\cal{L}}_{2}  = & {\cal{L}}_{1}
\end{aligned}
\end{equation*}

\begin{equation*}
\label{durative-refinement}
\begin{aligned}
T_2 = & \ (T_1 - \{t\}) \cup \{ v_t^s, v_t^e\}  \\
\alpha_2 = & \alpha_1 \cup \{(v_t^s, \tstart(a));(v_t^e, \tend(a))\} \\
{\cal{C}}_{2} = & \ {\cal{C}}_{1} \cup \{ v^{s}_{t} < v^{e}_{t}, v^{e}_{t} - v^{s}_{t} = d\} \\
{\cal{L}}_{2}  = & {\cal{L}}_{1} \cup \{ l \ | \ \forall p \in \tinv(a) \ l = \langle \tstart(a) \overset{p}{\rightarrow} \tend(a)  \rangle \} \\
\end{aligned}
\end{equation*}

\begin{figure}
    \centering
    \includegraphics[scale= 0.45]{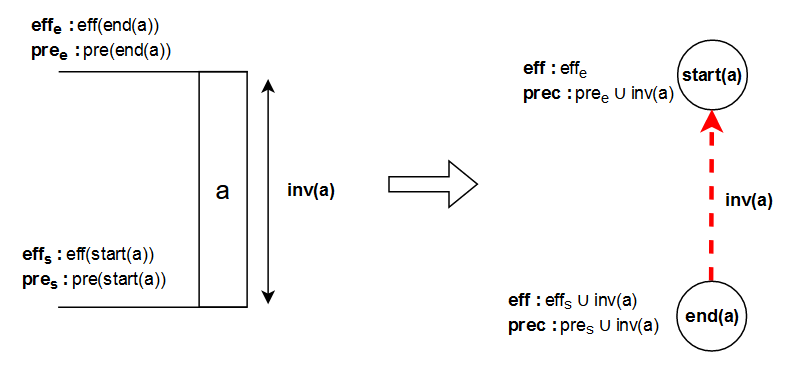}
    \caption{Our compilation of durative actions into snap actions. The invariant conditions $\tinv(a)$ are added to the preconditions of $start(a)$ and $end(a)$ and a causal link protecting $inv(a)$ (represented as a red dashed line) is added between the two snap actions.}
    \label{compiling}
\end{figure}

\section{Hierarchical Temporal Planning Event}
\label{temporalEventPlanning}

The particularity of our approach is that it works by interleaving two steps to take advantage of the heuristics developed for non-temporal hierarchical planning. The first step is to compile a temporal problem in a non temporal one. To that end, we refine abstract tasks and durative actions into instantaneous non temporal actions called \textit{snap actions}, and we search for a solution plan by guiding this search with non temporal POCL HTN heuristics, and by {\it checking} constraint consistency. The second step is to {\it find} time assignments matching the temporal constraints by using a simple CSP solver.

In this section, we present the search procedure implemented in our approach, called HTEP (Hierarchical Temporal Event Planner). We first give an overview of its search procedure based on hybrid planning \cite{bercher_hybrid}, then we detail the particular way of handling specific temporal flaws of our approach, and we terminate by presenting the implemented heuristics.

\subsection{Search Procedure}

The HTEP search procedure is given in Algorithm~\ref{algoTEP}. It takes as input a hierarchical temporal planning problem and returns a temporal partial plan. The procedure starts by compiling the tasks of the initial plan $\pi_0$ into snap actions (line 1). Recall that $\pi_0$ contains two special snap actions: $t_0$ with no precondition but with $s_0$ as effects and $t_\infty$ with the goal $g$ as precondition but no effects and $v^{e}_{t_0} < v^{s}_{t_\infty}$ in ${\cal{C}}$. Then, $\pi_0$  is added to the pending list of partial plans to explore $open$ (line 2) and the main loop starts (line 3). At each iteration a plan $\pi$ is non-deterministically selected in $open$ (line 4). Then, the flaws of $\pi$ are computed (line 5). If the plan has no flaws (line 6), HTEP searches an assignment of the time variables of $\pi$ by using a CSP that matches its constraints. If such assignment exists (line 8), $\pi$ is solution. Otherwise, one flaw is deterministically selected (line 9). Solving this flaw, generates a new set of partial temporal plans,  which are added to the $open$ list (line 10).

\begin{algorithm}[!t]
\SetAlgoLined
\caption{{\sf HTEP}(${\cal{S}}, {\cal{T}}, {\cal{A}}, {\cal{M}}, {s_0}, \pi_0, g)$}
\label{algoTEP}
\DontPrintSemicolon
\SetKwFunction{TEP}{\sf TEP}
\SetKwData{open}{open}
\SetKwData{flaws}{flaws}
\SetKwData{failure}{Failure}

$\open \leftarrow \{\pi_0$\}\;
\While{$\open \ne \emptyset$}{
    $\pi \leftarrow \text{non-deterministically select plan in }open$\;
    $\flaws \leftarrow \text{the set of flaws of }\pi$\;
    \If{$\flaws = \emptyset$}{
        $V \leftarrow \text{search of time variable assignment of $\pi$}$\;
        \lIf{$V \neq \emptyset$}{
            \textbf{return }$\pi$ and $V$
        }
    }
    $\phi \leftarrow \text{deterministically select a flaw in }\flaws$\;
    $\open \leftarrow \open \cup solveFlaw(\omega, \phi)$\;
}
\Return \failure;
\end{algorithm}

\subsection{Temporal Flaws}
\begin{figure*}[!th]
     \begin{subfigure}[b]{0.49\textwidth}
         \centering
         \includegraphics[width=0.5\textwidth]{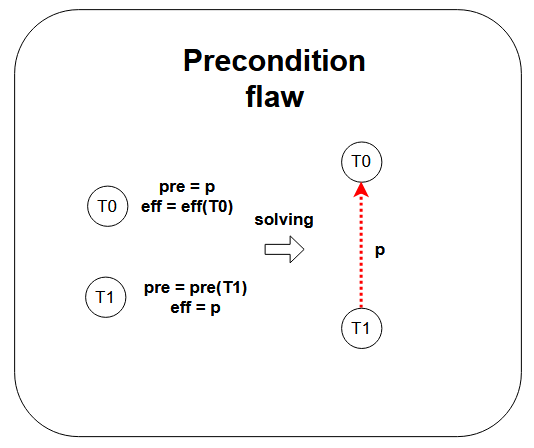}
        \caption{A Precondition flaw is solved by adding a \textit{causal link} from a provider event and the needer event.}
                 \label{preconditionFlaw}
     \end{subfigure}
     \begin{subfigure}[b]{0.49\textwidth}
         \centering
         \includegraphics[width=0.6\textwidth]{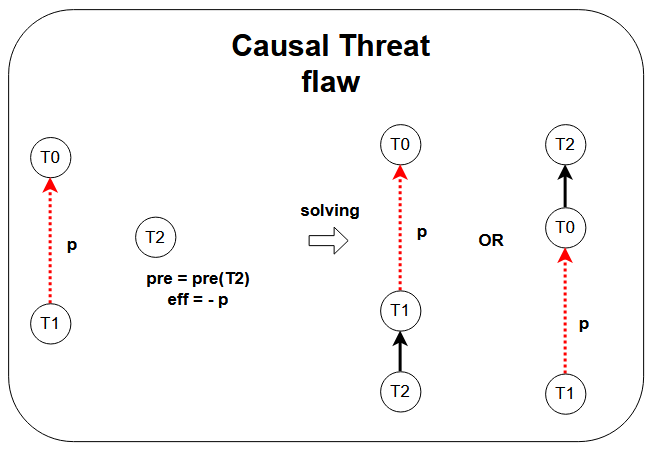}
        \caption{A Causal Threat flaw is solved by ordering the threatening event either before or after the causally linked events.}
                 \label{causalThreatFlaw}
     \end{subfigure}
    \begin{subfigure}[b]{0.49\textwidth}
         \centering
         \includegraphics[width=0.5\textwidth]{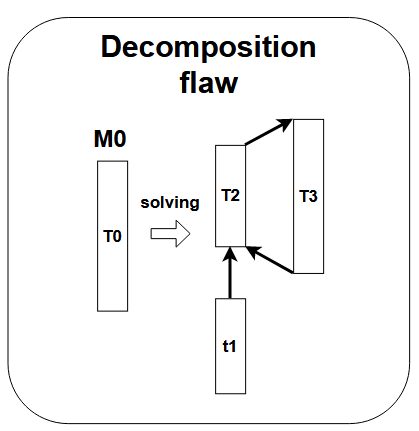}
             \caption{Decomposition flaws are solved by applying a method, temporal constraints are applied over start and end point of tasks.}
             \label{decompositionFlaw}
     \end{subfigure}
         \begin{subfigure}[b]{0.49\textwidth}
         \centering
         \includegraphics[width=0.55\textwidth]{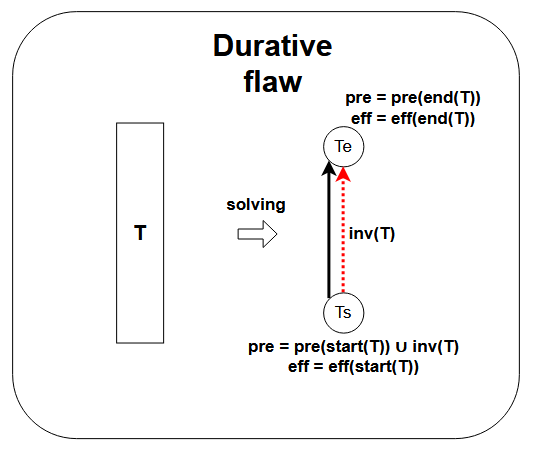}
             \caption{Durative flaws are solved by compiling a durative task into snap actions.}
             \label{durativeFlaw}
     \end{subfigure}
     \caption{The four types of flaws encountered by our planner and how to solve them. On every figure, black arrows represent ordering constraints and red dashed arrows the causal links.} \label{flaws}
\end{figure*}

In addition to the classic flaws in POCL (precondition and causal threat flaws), HTEP requires the management of two specific types of flaws: the decomposition flaws that are repaired by applying a method and by decomposing an abstract temporal task in primitive temporal tasks, and the durative flaws that can be repaired by decomposing a durative action into snap actions. We detail each flaw encountered in the following:


\begin{itemize}
    \item \textbf{Precondition flaw: }This flaw occurs for each precondition introduced in the partial plan. It is resolved by adding a \textit{causal link} between a preceding snap action having as effect the required precondition. We denote a causal link $c$ by $: t_1 \xrightarrow{p} t_2$. The resolution of this flaw is represented on Figure~\ref{preconditionFlaw}.
    \item \textbf{Causal threat flaw:} There is a causal threat flaw for each causal link $c : t_1 \xrightarrow{p} t_2$ and each snap action $a$ (with $p \in del(a)$) that is neither ordered before $t_1$ nor ordered after $t_2$. A causal threat is resolved either by adding a constraint: $a < t_1$ or $t_2 < a$ as represented on Figure~\ref{causalThreatFlaw}.
    \item \textbf{Decomposition flaw:} As in hybrid planning, decomposition flaw occurs for each abstract task still in the partial plan. They are solved by decomposing the abstract task with a method. Note that the decomposed tasks are still considered as durative and still need to be compiled into snap actions. This flaw is represented on Figure~\ref{decompositionFlaw}.
    \item \textbf{Durative flaw:} A durative flaw occurs for each durative action still in the plan. In order to solve this flaw, the durative action is decomposed into two snap actions representing the start and the end points of the task. Finally, a causal link protecting the invariant preconditions is added between the two snap actions. These invariant preconditions are then added as preconditions of the start snap action.

\end{itemize}

\subsection{Partial Order Planning heuristics}

The HTEP search procedure relies on two selection functions. The first one performs a non deterministic choice over the set of partial temporal plans in the $open$ list, and decides which partial plan to explore first (line 4). This function is called \textit{plan selection heuristic} and greatly impact both the search performances (e.g. the time required to find a solution plan) but also the quality of the returned plan (e.g. the cost of the plan according to some optimization function). The second selection function, called \textit{flaw selection heuristic}, selects (line 9) the flaw to be solved in the current partial plan to explore. Note that every flaw in the partial plan will eventually have to be solved in order to find a solution network. Hence, the \textit{admissibility} of a POCL procedure only depends on the \textit{plan selection} heuristic. However, the \textit{order} in which the flaws are resolved, defined by the flaw selection heuristic, greatly impacts the search performances of the procedure. In the following we will present the plan selection and flaw heuristics implemented in HTEP.

\subsubsection{Plan selection heuristics}

In the literature, there are two main categories of plan selection heuristics. The first type of heuristics has a POCL related approach which analyzes the flaws of a partial plan to infer a heuristic value. A well-known heuristics has been proposed in \cite{nguyen01} and simply counts the number of open conditions to satisfy in the partial plan. This idea has been further refined in PANDA~\cite{bercher_hybrid} where the use of TDG (Task Decomposition Graphs) allows to also estimate the number of open conditions that will be introduced by refining the plan. This led to two plan selection heuristics. The first one denoted $h_{\textit{TC}}$ computes the cardinality of the mandatory tasks that will appear in the partial plan decomposition. It is usually summed with the number of flaws remaining in the plan forming a heuristics denoted $h_{\textit{\#F + TC}}$. The second TDG heuristic is denoted $h_{\textit{MME}}$ and estimates the number of modifications required to refine the partial plan into a solution one. The $h_{\textit{MME}}$ has been further refined to take into account the number of causal links already introduced into the plan by a heuristic denoted $h_{\textit{TDGm}}$. Finally, FAPE \cite{fape} has proposed a plan selection heuristics in their chronicle planner. This heuristics is also an estimation of the remaining effort required to obtain a solution from a partial plan. In the following we will denote this heuristic $h_{\textit{FAPE}}$.

The second type of heuristics adapt heuristics from non hierarchical planning. The two most notable approaches have been proposed in \cite{nguyen01} where an adaptation of the Fast Forward heuristic \cite{ff} has been proposed. In addition, the ADD heuristics \cite{hsp} has been adapted in the VHPOP planner \cite{Younes03}. These heuristics are very well suited for non-hierachical POCL planning with task insertion, which is not considered here.

In that regard, we decided to use the two TDG plan selection heuristics $h_{\textit{\#F + TC}}$ and $h_{\textit{TDGm}}$ used in PANDA and the plan selection heuristic presented in FAPE for our experimentation.

\subsubsection{Flaw selection heuristics}

When it comes to flaw selection heuristics, the known strategies aim at reducing the branching factor of the search space by resolving the flaws with the fewest number of resolvers first. It is usually expressed as a priority list to follow. To our knowledge, there is no recent comparison between the current flaw selection heuristics. In that regard, we have chosen to implement in HTEP the flaw selection heuristics of PANDA \cite{bercher_hybrid} (which is the LCFR heuristics presented in \cite{lcfr}) and the priority list presented in FAPE \cite{fape}. Note that the heuristic used in FAPE is a refinement of the LCFR heuristic: while LCFR prioritizes the flaws with the fewest resolvers, the FAPE flaw selection heuristics also prioritizes unrefined tasks and preconditions first.

\section{Experimentation}
\label{experimentation}

\begin{figure*}[!t]
     \centering
     \begin{subfigure}[b]{0.49\textwidth}
         \centering
         \includegraphics[width=0.8\textwidth]{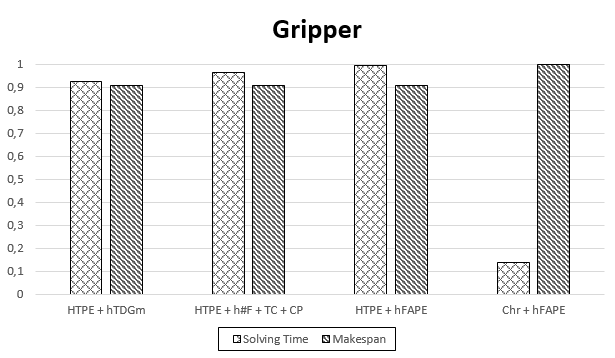}
         \label{satelliteTO}
     \end{subfigure}
     \begin{subfigure}[b]{0.49\textwidth}
         \centering
         \includegraphics[width=0.8\textwidth]{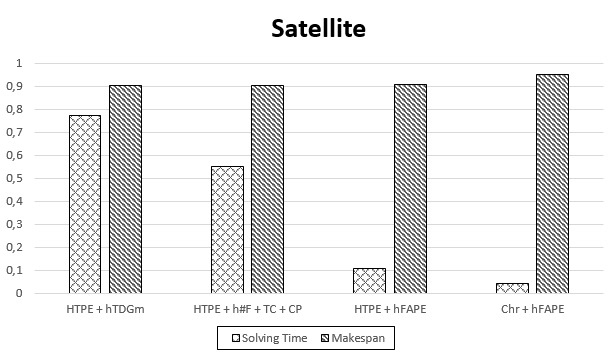}
         \label{satellitePO}
     \end{subfigure}
    \begin{subfigure}[b]{0.49\textwidth}
         \centering
         \includegraphics[width=0.8\textwidth]{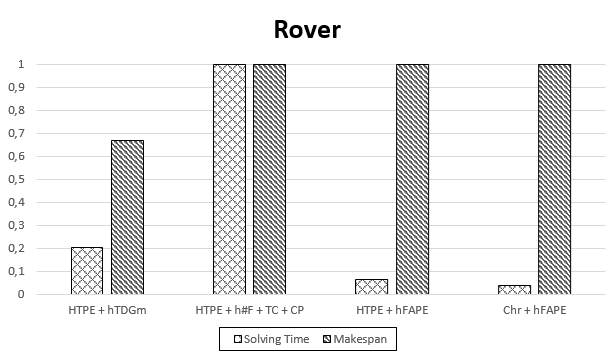}
         \label{miconicTO}
     \end{subfigure}
         \begin{subfigure}[b]{0.49\textwidth}
         \centering
         \includegraphics[width=0.8\textwidth]{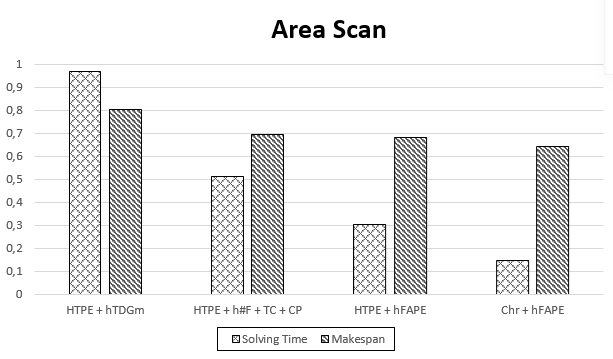}
         \label{miconicTO}
     \end{subfigure}
        \caption{Results for the five configurations, with the IPC metric. Each diagram represents the results on one domain.}
        \label{results}
\end{figure*}


In this section, we will compare our \textit{Temporal Event} approach with a \textit{Chronicle} approach. Both approaches have been been coded and tested on the same device. We have used as reference the chronicle planner described in \cite{fape} as it is the current state-of-the-art of hierarchical temporal planning. As both algorithms use a CSP solver in their procedures, we will use the same CSP solver as well.
All the benchmarks and code used to produce these results will be freely available for result reproduction.

\subsection{Experimental Setup}

We will compare the results on three indicators:
\begin{itemize}
    \item \textbf{Solving time}: it represents the time spent to solve the problem (from instantiation to solution)
    \item \textbf{Makespan of the solution}: it represents the overall length of the plan, meaning the time between the initial state and the final task end point.
\end{itemize}

The results will be presented by scoring these two indicators with the IPC scoring metric.

In this paper we consider four configurations, which are presented below:

\begin{itemize}

    \item \textbf{$HTEP + h_{\textit{TDGm}}$: }in this first configuration, HTEP is used with the $h_{\textit{TDGm}}$ heuristics described in \cite{TDGm} adapted to our temporal event representation.

    \item \textbf{$HTEP + h_{\textit{\#F + PC}}$: }in this configuration, the Temporal Event Planner is used with the first TDG heuristics proposed in \cite{bercher_hybrid} associated with $h_{\textit{\#F + PC}}$ heuristics.

    \item \textbf{$TE + h_{\textit{FAPE}}$: }in this configuration, HTEP is used with the plan selection heuristics used by the FAPE planner \cite{fape}.

    \item \textbf{$Chr + h_{\textit{FAPE}}$: }this is the chronicle planner encoded to mimic the behavior of FAPE \cite{fape}. It uses both the plan selection and flaw selection heuristics described in \cite{fape} and uses a chronicle representation.

\end{itemize}

Note that all configurations use the same Flaw Selection heuristics described in \cite{fape}.


We chose four temporal planning domains for our tests. These domains require different levels of concurrency to be solved. We categorized these problems according to the three Cushing concurrency categories \cite{cushing}:
  \begin{itemize}
    \item \textbf{Gripper: }this domain is a sequential one, it is the simplest domain with no temporal concurrency required. These problems are members of the first Cushing \cite{cushing} category where all solutions are sequential.
    \item \textbf{Satellite: }this domain is also a sequential one but optimization over the \textit{makespan} of the solution is possible (e.g. there are several sequential plans possible with different makespans). These problems are also members of the first Cushing category.
    \item \textbf{Rover: }in this domain, concurrency is possible although not required. The planner can either choose to find a simple non concurrent plan or a more efficient concurrent one. These problems are members of the second Cushing category where solutions can either be sequential or concurrent.
    \item \textbf{Area scan: }this domain models a set of heterogeneous devices aiming at cooperating in order to scan areas. The devices can either cooperate to reach their destination faster or act by their own, resulting in less effective plans. It presents both sequential and concurrent solution plans, with numerous makespan optimization possible.
\end{itemize}

We expressed all domains and problems in HDDL 2.1 language proposed by \cite{HDDL21}. All experiments were run on a single core of a Intel Core i7-9850H CPU, with a limit of 8GB of RAM over 600 seconds. The code and benchmarks will be made available if this paper is accepted.

\subsection{Results}

The results are presented on Figure \ref{results}.

We can see that there are some tendencies over all domains. On all domains and for both heuristics, the Temporal Event planner outperforms its chronicle counterpart with regards to the \textit{Solving time} metric. HTEP handles non temporal constraints and flaws which leads to simpler plan refinements and flaws compared to a Chronicle planner. In our opinion, this is what explains the discrepancies between the $HTEP + h_{\textit{FAPE}}$ and $Chr + h_{\textit{FAPE}}$ configurations. In addition, the HTEP approach benefits from the use of classical HTN heuristics: the $HTEP + h_{\textit{TDGm}}$ and $HTEP + h_{\textit{\#F + PC}}$ configurations are the best performing ones on all domains. We can notice that the $h_{\textit{\#F + PC}}$ seems to be the best performing one on the Rover domain while $h_{\textit{TDGm}}$ is more efficient on the other domains. As explained in \cite{bercher_hybrid}, the performance of each heuristics depends on the domain definition of the problem and a heuristics can be more or less informative depending on the domain.
As a Chronicle approach should handles a larger set of constraints it should be able to eliminate partially refined plan sooner in the refinement process than HTEP. However this case does not happen often as many HTN domains describe the necessary temporal ordering through methods and task decomposition. The main sources of partial plan elimination are unsolvable non temporal flaws.
Overall, it seems that the $HTEP + h_{\textit{TDGm}}$ configuration is the most efficient one when it comes to the \textit{Solving Time} metric.

Concerning the \textit{Makespan}, we can see that the $Chr + h_{\textit{FAPE}}$ configuration is the most efficient one. As this configuration handles the full temporal constraints, it can prioritize the best one in terms of makespan when two have equal heuristic value. This leads to higher quality plans in most domains. On the other hands, the HTEP configurations remain competitive with the Chronicle approach. The exception to this is displayed on the Rover domain where the $HTEP + h_{\textit{TDGm}}$ configuration is the lowest one makespan wise. This heuristics aims at finding the solution with the fewest number of refinement required, regardless of the solution plan makespan. Note that if $h_{\textit{TDG}}$ is the best performing one on the Area Scan domain is due to the fact that some instances have not been solved by others configurations, thus increasing its score.

Overall, the configuration combining HTEP and the classical HTN heuristics seems to outperform the \textit{Chronicle} approaches. This formalism allows for simpler constraints representation and managements. It also benefits from the hybrid planning heuristics and the HTN formalism which often provides the necessary ordering constraints to the planner.

\section{Discussion}

All along this paper, we used a compilation of temporal actions into snap actions by representing the invariant condition of a temporal action through a POCL causal link. The invariant are treated as precondition of the start snap action and protected until the end snap task through this causal link (see Figure \ref{compiling}). This compilation implies a that HTEP can not solve temporal in the third Cushing's category: in temporal planning, invariant condition can not be seen as precondition for the whole temporal action. Instead, they should be seen as \textit{postconditions} of the start snap action, meaning that invariant condition should be verified right \textit{after} the execution of the start. This subtlety actually adds a new layer of complexity in planning as it allows to define \textit{necessary concurrent actions}. This has been demonstrated and explained by Cushing in his three temporal hierarchical problem classes definition \cite{cushing}.


\section{Conclusion}
\label{conclusion}

In this paper, we have presented an approach to represent and solve Temporal HTN problems by using Temporal Events. This approach relaxes the temporal problem in a simpler one, which allows to apply classical HTN search heuristics to it. We have compared this approach with the Chronicle one, which is the current state of the art in hierarchical temporal planning. We have shown that the Temporal Event approach outperforms it in terms of time spent to find a solution and is comparable to it when it comes to the quality of the solution plans. HTEP can still be improved by applying other classical HTN search techniques. In addition, we want to improve the compilation made in HTEP in order to allow it to solve temporal problems in the third Cushing category.


\begin{thebibliography}{24}
\providecommand{\natexlab}[1]{#1}

\bibitem[{Abdulaziz and Koller(2022)}]{abdulaziz22}
Abdulaziz, M.; and Koller, L. 2022.
\newblock Formal Semantics and Formally Verified Validation for Temporal
  Planning.
\newblock In \emph{{AAAI} Conference on Artificial Intelligence}, 9635--9643.

\bibitem[{Asuncion et~al.(2005)Asuncion, Castillo, Fdez-Olivares, Garcia-Perez,
  Gonzalez-Munoz, and Palao}]{siadex}
Asuncion, M.; Castillo, L.; Fdez-Olivares, J.; Garcia-Perez, O.;
  Gonzalez-Munoz, A.; and Palao, F. 2005.
\newblock {SIADEX}: An interactive knowledge-based planner for decision support
  in forest fire fighting.
\newblock \emph{AI Commun.}, 18: 257--268.

\bibitem[{Au et~al.(2003)Au, Ilghami, Kuter, Murdock, Nau, Wu, and
  Yaman}]{shop2}
Au, T.-C.; Ilghami, O.; Kuter, U.; Murdock, J.~W.; Nau, D.~S.; Wu, D.; and
  Yaman, F. 2003.
\newblock SHOP2: An HTN Planning System.
\newblock \emph{J. of Artif. Intell. Res.}, 20: 379--404.

\bibitem[{Baier, Bacchus, and Mcilraith(2009)}]{hsp}
Baier, J.; Bacchus, F.; and Mcilraith, S. 2009.
\newblock A Heuristic Search Approach to Planning with Temporally Extended
  Preferences.
\newblock \emph{Artif. Intell.}, 173: 593--618.

\bibitem[{Barreiro et~al.(2012)Barreiro, Boyce, Do, Frank, Iatauro, Kichkaylo,
  Morris, Ong, Remolina, Smith et~al.}]{europa}
Barreiro, J.; Boyce, M.; Do, M.; Frank, J.; Iatauro, M.; Kichkaylo, T.; Morris,
  P.; Ong, J.; Remolina, E.; Smith, T.; et~al. 2012.
\newblock {EUROPA: A platform for AI planning, scheduling, constraint
  programming, and optimization}.
\newblock \emph{4th International Competition on Knowledge Engineering for
  Planning and Scheduling (ICKEPS)}.

\bibitem[{Bechon et~al.(2014)Bechon, Barbier, Infantes, Lesire, and
  Vidal}]{bechon14}
Bechon, P.; Barbier, M.; Infantes, G.; Lesire, C.; and Vidal, V. 2014.
\newblock {HiPOP: Hierarchical Partial-Order Planning}.
\newblock In \emph{Starting AI Researchers' Symposium}.

\bibitem[{Bercher et~al.(2017)Bercher, Behnke, Höller, and Biundo}]{TDGm}
Bercher, P.; Behnke, G.; Höller, D.; and Biundo, S. 2017.
\newblock {An Admissible HTN Planning Heuristic}.
\newblock In \emph{IJCAI}, 480--488.

\bibitem[{Bercher, Keen, and Biundo(2014)}]{bercher_hybrid}
Bercher, P.; Keen, S.; and Biundo, S. 2014.
\newblock Hybrid {Planning} {Heuristics} {Based} on {Task} {Decomposition}
  {Graphs}.
\newblock \emph{Proceedings of the International Symposium on Combinatorial
  Search}.

\bibitem[{Bit{-}Monnot et~al.(2020)Bit{-}Monnot, Ghallab, Ingrand, and
  Smith}]{fape}
Bit{-}Monnot, A.; Ghallab, M.; Ingrand, F.; and Smith, D.~E. 2020.
\newblock {FAPE}: a {Constraint}-based {Planner} for {Generative} and
  {Hierarchical} {Temporal} {Planning}.
\newblock \emph{CoRR}, 2010.13121.

\bibitem[{Broxvall and Jonsson(2003)}]{broxvall03}
Broxvall, M.; and Jonsson, P. 2003.
\newblock Point algebras for temporal reasoning: Algorithms and complexity.
\newblock \emph{Artif. Intell.}, 149(2): 179--220.

\bibitem[{Celorrio, Jonsson, and Palacios(2015)}]{jimenez15}
Celorrio, S.~J.; Jonsson, A.; and Palacios, H. 2015.
\newblock Temporal Planning With Required Concurrency Using Classical Planning.
\newblock In \emph{ICAPS}, 129--137.

\bibitem[{Cushing(2007)}]{cushing}
Cushing, W. 2007.
\newblock Evaluating {Temporal} {Planning} {Domains}.
\newblock \emph{ICAPS}, 105--112.

\bibitem[{Dechter, Meiri, and Pearl(1991)}]{dechter91}
Dechter, R.; Meiri, I.; and Pearl, J. 1991.
\newblock Temporal Constraint Networks.
\newblock \emph{Artif. Intell.}, 49(1-3): 61--95.

\bibitem[{Fox and Long(2003)}]{fox03}
Fox, M.; and Long, D. 2003.
\newblock {PDDL2.1:} An Extension to {PDDL} for Expressing Temporal Planning
  Domains.
\newblock \emph{J. of Artif. Intell. Res.}, 20: 61--124.

\bibitem[{Goldman(2006)}]{goldman06}
Goldman, R. 2006.
\newblock Durative Planning in HTNs.
\newblock In \emph{Proceedings of the International Conference on Automated
  Planning and Scheduling}, 382--385.

\bibitem[{Hoffmann and Nebel(2011)}]{ff}
Hoffmann, J.; and Nebel, B. 2011.
\newblock {The FF Planning System: Fast Plan Generation Through Heuristic
  Search}.
\newblock \emph{J. of Artif. Intell. Res.}, 14.

\bibitem[{H{\"{o}}ller et~al.(2020)H{\"{o}}ller, Behnke, Bercher, Biundo,
  Fiorino, Pellier, and Alford}]{holler20}
H{\"{o}}ller, D.; Behnke, G.; Bercher, P.; Biundo, S.; Fiorino, H.; Pellier,
  D.; and Alford, R. 2020.
\newblock {HDDL:} An Extension to {PDDL} for Expressing Hierarchical Planning
  Problems.
\newblock In \emph{{AAAI} Conference on Artificial Intelligence}, 9883--9891.

\bibitem[{Joslin and Pollack(1994)}]{lcfr}
Joslin, D.; and Pollack, M.~E. 1994.
\newblock {Least-Cost Flaw Repair: A Plan Refinement Strategy for Partial-Order
  Planning}.
\newblock In Hayes{-}Roth, B.; and Korf, R.~E., eds., \emph{NCAI}, 1004--1009.

\bibitem[{Lallement, de~Silva, and Alami(2018)}]{htnRobotics}
Lallement, R.; de~Silva, L.; and Alami, R. 2018.
\newblock {HATP: Hierarchical Agent-Based Task Planner}.
\newblock In \emph{AAMAS}, 1823--1825.

\bibitem[{Lemai(2004)}]{ixtet}
Lemai, S. 2004.
\newblock {IXTET}-{EXEC}: planning, plan repair and execution control with time
  and resource management.

\bibitem[{Milot et~al.(2021)Milot, Chauveau, Lacroix, and
  Lesire}]{htnTaskAuction}
Milot, A.; Chauveau, E.; Lacroix, S.; and Lesire, C. 2021.
\newblock {Solving Hierarchical Auctions with HTN Planning}.
\newblock In \emph{{ICAPS workshop on Hierarchical Planning}}.

\bibitem[{Nguyen and Kambhampati(2001)}]{nguyen01}
Nguyen, X.; and Kambhampati, S. 2001.
\newblock Reviving Partial Order Planning.
\newblock In \emph{IJCAI}, 459--464.

\bibitem[{Pellier et~al.(2023)Pellier, Albore, Fiorino, and
  Bailon-Ruiz}]{HDDL21}
Pellier, D.; Albore, A.; Fiorino, H.; and Bailon-Ruiz, R. 2023.
\newblock {HDDL 2.1: Towards Defining an HTN Formalism with Time}.
\newblock In \emph{6th ICAPS Workshop on Hierarchical Planning (HPlan 2023)}.

\bibitem[{Younes and Simmons(2003)}]{Younes03}
Younes, H. L.~S.; and Simmons, R.~G. 2003.
\newblock {VHPOP:} Versatile Heuristic Partial Order Planner.
\newblock \emph{J. of Artif. Intell. Res.}, 20: 405--430.

\end{thebibliography}
\end{document}